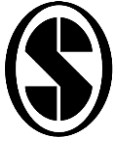

Corresponding Author:
Majid Khadiv, Center of Excellence in Robotics and Control, Advanced Robotics & Automated Systems (ARAS) Lab, Department of Mechanical Engineering, K. N. Toosi University of Technology, Tehran, Iran.
Email: mkhadiv@mail.kntu.ac.ir

# Online Adaptation for Humanoids Walking On Uncertain Surfaces

**Majid Khadiv[†], S. Ali. A. Moosavian[†], Aghil Yousefi-Koma[††], Hessam Maleki[††], Majid Sadedel[††]**

[†] Center of Excellence in Robotics and Control, Advanced Robotics & Automated Systems (ARAS) Lab, Department of Mechanical Engineering, K. N. Toosi University of Technology, Tehran, Iran.
[††] Center of Advanced Systems and Technologies (CAST), School of Mechanical Engineering, College of Engineering, University of Tehran, Tehran, Iran.

## Abstract

In this paper, an online adaptation algorithm for bipedal walking on uneven surfaces with height uncertainty is proposed. In order to generate walking patterns on flat terrains, the trajectories in the task space are planned to satisfy the dynamic balance and slippage avoidance constraints, and also to guarantee smooth landing of the swing foot. To ensure smooth landing of the swing foot on surfaces with height uncertainty, the preplanned trajectories in the task space should be adapted. The proposed adaptation algorithm consists of two stages. In the first stage, once the swing foot reaches its maximum height, the supervisory control is initiated until the touch is detected. After the detection, the trajectories in the task space are modified to guarantee smooth landing. In the second stage, this modification is preserved during the Double Support Phase (DSP), and released in the next Single Support Phase (SSP). Effectiveness of the proposed online adaptation algorithm is experimentally verified through realization of the walking patterns on the SURENA III humanoid robot, designed and fabricated at CAST. The walking is tested on a surface with various flat obstacles, where the swing foot is prone to either land on the ground soon or late.

## Keywords:

Humanoid Robot, Biped robot, Gait planning, Online adaptation.

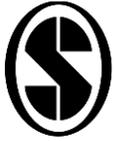

## 1. Introduction

The most significant advantage of legged robots over wheeled robots is their ability to deal with uneven terrains. This ability would be more highlighted when one considers that the environment around us has been constructed for human beings. As a result, exploiting this advantage to make the biped robots able to walk on various uneven terrains would be a step toward realizing the dream of employing humanoid robots as assistants and servants in our daily life. Regarding this matter, investigating strategies for online adaptation of humanoid robots, walking on unknown uneven surfaces, would be of great significance.

The procedure of gait planning for humanoid robots usually is composed of generating task space trajectories and adopting inverse kinematics to obtain consistent joint space trajectories[1-4]. The Task space for the lower-body of humanoid robots includes the feet and the pelvis. After determining gait parameters in the high-level control unit of the robot, the trajectories for the feet are planned consistent with the surface to preserve smooth landing of the swing foot. Then, the pelvis trajectory is generated to guarantee the dynamic balance and slippage avoidance. The most well-known criterion for the dynamic balance is certainly the Zero Moment Point (ZMP) which has been introduced by Vukobratovic et al.[5, 6]. In the case of using ZMP to generate a feasible pelvis trajectory, two approaches are typically adopted. In the first approach that is suitable for real-time trajectory generation, a simple model based on the Linear Inverted Pendulum Mode (LIPM)[7] is exploited[2, 3, 8, 9]. The other approach is based on thorough information of the robot links, and is an offline method due to its huge computation burden[10, 11]. Then, generated trajectories in the task space are projected into the joint space to yield desired joint trajectories, using inverse kinematics[12]. Finally, the low-level control system is responsible for tracking the obtained desired trajectories[13-15]. Due to the presence of uncertainties, the preplanned gaits should be modified in real-time to be implemented in real environment. These uncertainties may exist because of the robot kinematics and dynamics modeling errors as well as the environment map estimation. Hence, in order to realize the generated walking pattern, some online modification strategies should be exploited to cope with these uncertainties. The research studies that have been carried out to address this problem employ a supervisory control paradigm to deal with each uncertainty. In this notion, Huang et al. introduced a reflex control based on ankle and hip strategies and foot landing height modifications[16]. Kim et al. adopted six strategies to deal with unexpected motions that are generated due to uncertainties, in their online control topology[17]. Kajita et al. used task space modification to be a general algorithm for controlling bipedal robots with different structures[18]. Some other algorithms based on ankle and hip strategies were also suggested[19, 20].

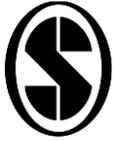

Though some amendments are considered during gait planning to realize landing without impact or slip on various ground surfaces based on an environment map, local irregularities on the ground surface may cause soon or late landing of the swing foot which makes the robot vulnerable to lose its balance[16, 17]. As a result, in order to tackle with this problem, it is crucial to monitor the swing foot state with respect to the ground surface. In this notion, some researchers employed force/torques sensors embedded to the ankle of the robot to examine state of the contact between the swing foot and the ground surface. In this method, usually a threshold is considered to distinguish the contact/non-contact states. In this context, Morisawa et. al. introduced their algorithm based on two factors: considering suitable desired impedance gains during different phases of walking, and updating the landing position as a new reference[21]. They employed an admittance approach for force control and a damping algorithm for control of the interacting moments. Also, Son et al. constructed their adaptation algorithm based on alleviating the interacting force in the case that the swing foot impacts the ground[22]. They implemented their algorithm on the basis of variable desired interacting force in a simulation environment. Nishiwaki et al. adopted a damping method for controlling the interacting forces and torques[23]. Similarly, some other algorithms based on compliance of the landing foot and reducing position gains for the swing foot have been proposed[24, 25].

Employing force/torque sensors to detect the contact state may cause large impacts during fast collision of the swing foot to the ground surface. To cope with this problem, Sato et. al. exploited flexible foot mechanism and contact sensors to adapt the walking with various irregularities on the ground surface[26]. Also, Hashimoto et. al. used optical sensors at three corners of their triangular robot feet to detect the contact[27]. In their algorithm, modifications are applied on the vertical position trajectory as well as pitch and roll orientations. A similar sensing mechanism with four optical sensors had been investigated, by Kang et. al.[28]. However, in this article, our main goal is to exploit contact switch sensors embedded to the soles of a humanoid robot, in order to realize smooth landing of the swing foot in the case of soon or late landing. By adjusting these sensors in a certain distance from the sole, contact between the swing foot and the ground surface may be detected in advance. After the detection, the swing foot trajectory is modified to realize landing without impact or slippage. The proposed strategy is based on the swing foot trajectory modification in the case of soon landing, and both swing foot and pelvis trajectory modification in the case of late landing.

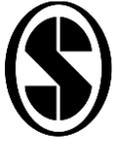

The rest of this paper is organized as follows: In section (2), a method for gait planning based on a comprehensive model of the robot is presented. Section (3) describes the proposed approach for online adaptation of the swing foot. The obtained results from the experimental implementation are discussed in Section (4). Finally, section (5) is dedicated to conclusions.

## 2. Gait planning

The humanoid robot SURENA III (Figure 1) is composed of 12 DOF in its lower-body and 19 DOF in its upper-body. This robot has 6 DOF in each leg ( 3 DOF in the hip, 1 DOF in the knee, 2 DOF in the ankle), 7 DOF in each arm ( 3 DOF in the shoulder, 1 DOF in the elbow, 3 in the wrist), 1 DOF in each hand (a simple gripper), one DOF in the torso and 2 DOF in the neck. In order to investigate bipedal walking of this robot, upper-body joints are considered to be fixed and the whole upper-body is modeled with a single rigid body. Actuation of the lower-body is done by EC motors. For power transmission, a combination of belt and pulley, and harmonic drive is used. The sensor layout includes incremental and absolute encoders at the motor output and gearbox output of each joint, one six axes force/torque sensors embedded to each ankle, an IMU mounted on the upper-body, and contact switches embedded to each sole.

In this section, a method for walking pattern generation in the task space (Figure 1.b) for walking on flat surfaces is described[4]. In this method, first the feet trajectories are generated respecting the constraints. Then, the pelvis trajectory is generated employing six parameters which parameterize trajectories in each direction.

### *2.1 Foot trajectory*

The topology that is adopted in this article for walking considers the feet parallel to the ground surface during walking, without heel-off or toe-off phases[29, 30]. Therefore, orientation of the feet remains constant, parallel to the ground surface, in all phases of the motion. In this topology, during the DSP, both feet are completely in contact with the ground surface. Furthermore, During the SSP, the stance foot is fixed to the ground surface. However, for the swing foot, the trajectories should be designed to realize a feasible walking pattern.

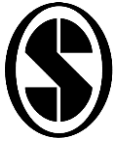
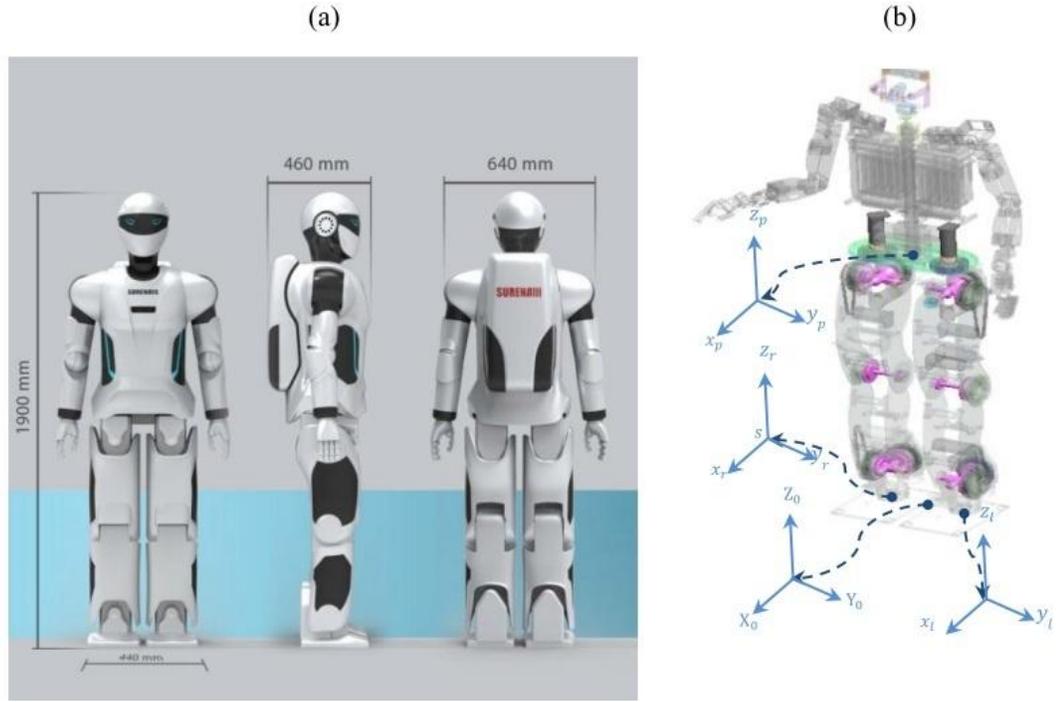

Figure 1. SURENA III, a humanoid robot designed and fabricated in CAST, University of Tehran, (a) three views of the robot whole body and its general attributes, (b) Developed model of the robot.

The swing foot in the direction of motion (x-direction) should move one stride (two steps), during the SSP period (Figure 2.a). In order to preserve slippage avoidance of the swing foot at landing instant, the velocity and acceleration at the end of the SSP are considered to be zero. Moreover, the trajectories in the task space should be $C^2$ continuous to preserve continuity of the ZMP trajectory, and also to yield smooth trajectories in the joint space. Hence, a fifth order polynomial is exploited in x direction to satisfy all the constraints:

$$\begin{cases} x_f(0) = 0 \\ \dot{x}_f(0) = 0 \\ \ddot{x}_f(0) = 0 \\ x_f(T_s) = 2D_s \\ \dot{x}_f(T_s) = 0 \\ \ddot{x}_f(T_s) = 0 \end{cases} \Rightarrow x_f = \sum_{i=0}^{5} a_i t^i \quad , \quad 0 \leq t \leq T_s \qquad (1)$$

where index $f$ stands for foot, and $D_s$ and $T_s$ specify the step length and the SSP period, respectively. Furthermore, $a_i$'s are coefficients that are computed consistent with the constraints, and $t$ indicates time.

In order to prevent self-collision of the legs during walking, trajectory of the swing foot in lateral direction (y-direction) is considered such that remains in a fixed distance with respect to the stance foot. This distance is considered equal to $L_p$ which is the distance between the feet centers, when the robot stands in an upright posture (Figure 2.b).

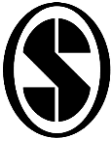

For the swing foot in the vertical direction (z direction), the height of the foot increases to a certain level (maximum height of the step), and then decreases to zero. In order to generate $C^2$ continuous trajectories, and prevent impact at landing, the velocity and acceleration at the start and end of the swing phase should be zero. Therefore, the constraints as well as the considered polynomial during the swing phase may be stated as:

$$\begin{cases} z_f(0) = 0 \\ \dot{z}_f(0) = 0 \\ \ddot{z}_f(0) = 0 \\ z_f(T_s/2) = H_{max} \\ z_f(T_s) = 0 \\ \dot{z}_f(T_s) = 0 \\ \ddot{z}_f(T_s) = 0 \end{cases} \Rightarrow z_f = \sum_{i=0}^{6} a_i t^i \quad , \quad 0 \le t \le T_s \tag{2}$$

In this set of equations, $z_f$ is the vertical trajectory of the swing foot, while $H_{max}$ represents the maximum height of the swing foot (Figure 2.b).

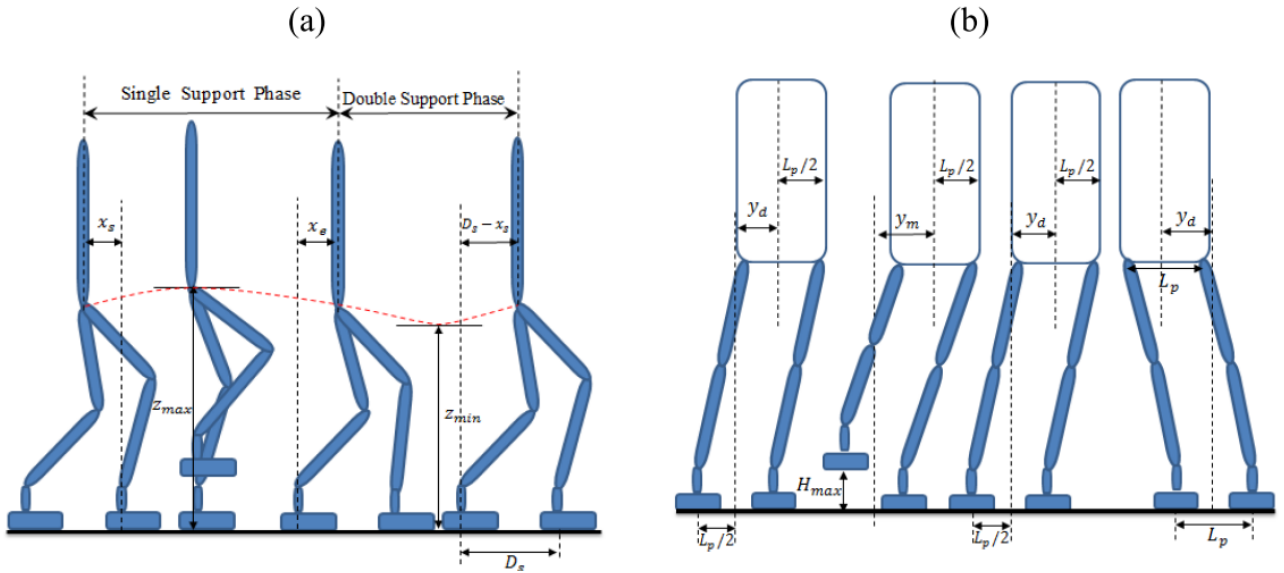

Figure 2. The gait parameters are specified on a model of the robot; (a) lateral view, (b) frontal view

## 2.2 Pelvis trajectory

After generating the feet trajectories for walking on flat surfaces, the pelvis trajectory should be designed to preserve the motion feasibility constraints. In fact, among infinitely many number of trajectories, those which satisfy the dynamic balance and slippage avoidance are feasible.

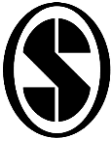

Since there are no pitch and roll joints in the upper-body structure of the robot, orientation of the upper-body is considered to be constant during walking. In fact, like human beings who do not exploit upper-body inclination for preserving dynamic balance during usual walking, in our gait planning method the upper-body remains in an upright posture during walking. Therefore, the three components of the pelvis orientation are set to zero. However, suitable trajectories for the three components of the pelvis reference point position (center of the pelvis) should be designed.

For the pelvis in x direction, two parameters are adopted which specify boundary conditions of the pelvis trajectory in the direction of motion. These parameters are $x_s$ which specifies sagittal distance between the pelvis center and the stance foot center at the start of the SSP, and $x_e$ which stands for the sagittal distance between the center of pelvis and the center of stance foot at the end of the SSP (Figure 2.a). Using these parameters as the boundary conditions and considering the velocity and acceleration continuity conditions, the pelvis trajectory in x direction can be written as:

$$\begin{cases} \begin{cases} x_{P,SSP}(0) = -x_s \\ x_{P,SSP}(T_s) = x_e \\ \dot{x}_{P,SSP}(0) = \dot{x}_{P,DSP}(T_c) \\ \ddot{x}_{P,SSP}(0) = \ddot{x}_{P,DSP}(T_c) \end{cases} \Rightarrow x_{P,SSP} = \sum_{i=0}^{3} a_i t^i \quad , \quad 0 \leq t \leq T_s \\ \begin{cases} x_{P,DSP}(T_s) = x_e \\ x_{P,DSP}(T_c) = D_s - x_s \\ \dot{x}_{P,DSP}(T_s) = \dot{x}_{P,SSP}(T_s) \\ \ddot{x}_{P,DSP}(T_s) = \ddot{x}_{P,SSP}(T_s) \end{cases} \Rightarrow x_{P,DSP} = \sum_{i=0}^{3} b_i t^i \quad , \quad T_s \leq t \leq T_c \end{cases} \tag{3}$$

where $x_{P,SSP}$ and $x_{P,DSP}$ specify the position in the direction of motion during the SSP and the DSP, respectively. Moreover, $T_s$ and $T_c$ are the SSP period and the step period, and $a_i$'s and $b_i$'s are constants that may be computed using the specified boundary conditions.

Similarly, for the pelvis in the lateral direction, two parameters are employed which specify the boundary conditions. These parameters are $y_d$ which describes the lateral distance between the pelvis center and the stance foot center at the start of SSP, and $y_m$ which specifies the maximum lateral distance between the pelvis center and the stance foot center (Figure 2.b). Considering these parameters as the boundary conditions, the $C^2$ continuous trajectory for the pelvis in the lateral direction is specified as:

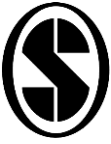

$$\begin{cases} \begin{cases} y_{P,SSP}(0) = y_d \\ y_{P,SSP}(T_s/2) = y_m \\ y_{P,SSP}(T_s) = y_d \\ \dot{y}_{P,SSP}(0) = -\dot{y}_{P,DSP}(T_c) \\ \ddot{y}_{P,SSP}(0) = -\ddot{y}_{P,DSP}(T_c) \end{cases} \Rightarrow y_{P,SSP} = \sum_{i=0}^{4} a_i t^i \quad , \quad 0 \le t \le T_s \\ \begin{cases} y_{P,DSP}(T_s) = y_d \\ y_{P,DSP}(T_c) = -y_d \\ \dot{y}_{P,DSP}(T_s) = \dot{y}_{P,SSP}(T_s) \\ \ddot{y}_{P,DSP}(T_s) = \ddot{y}_{P,SSP}(T_s) \end{cases} \Rightarrow y_{P,DSP} = \sum_{i=0}^{3} b_i t^i \quad , \quad T_s \le t \le T_c \end{cases} \quad (4)$$

Where $y_{P,SSP}$ and $y_{P,DSP}$ specify the pelvis trajectory in the lateral direction during the SSP and the DSP, respectively.

Finally, the parameters that describe the boundary conditions of the pelvis in the vertical direction are $z_{max}$ (the maximum pelvis height), and $z_{min}$ (the minimum pelvis height) which are illustrated in Figure 2.a. Hence, the pelvis trajectory in Z direction is:

$$\begin{cases} \begin{cases} z_{P1}(T_S/2) = z_{max} \\ z_{P1}(T_s + T_d/2) = z_{min} \\ \dot{z}_{P1}(T_S/2) = \dot{z}_{P2}(T_c + T_s/2) \\ \ddot{z}_{P1}(T_S/2) = \ddot{z}_{P2}(T_c + T_s/2) \end{cases} \Rightarrow z_{P1} = \sum_{i=0}^{3} a_i t^i \quad , \quad \frac{T_S}{2} \le t \le T_s + \frac{T_d}{2} \\ \begin{cases} z_{P2}(T_s + T_d/2) = z_{min} \\ z_{P2}(T_c + T_s/2) = z_{max} \\ \dot{z}_{P2}(T_s + T_d/2) = \dot{z}_{P1}(T_s + T_d/2) \\ \ddot{z}_{P2}(T_s + T_d/2) = \ddot{z}_{P1}(T_s + T_d/2) \end{cases} \Rightarrow z_{P2} = \sum_{i=0}^{3} b_i t^i \quad , \quad T_s + \frac{T_d}{2} \le t \le T_c + \frac{T_s}{2} \end{cases} \quad (5)$$

Where $z_{P1}$ is the pelvis trajectory in the lateral direction from the midpoint of the SSP to the midpoint of the DSP, while $z_{P2}$ is the vertical pelvis trajectory from the midpoint of the DSP to the midpoint of the SSP.

By changing the gait parameters that have been described, different walking patterns may be obtained. However, the necessary conditions for selecting these parameters are that they should satisfy both the dynamic balance (ZMP should be within the support polygon) and the slippage avoidance constraints (the required friction coefficient should be less than the available friction)[31]. Furthermore, the required torque and velocity in the joint space should be consistent with the actuators specifications. Finally, since the trajectory generation procedure is carried out in the task space, singularity avoidance should be carried out. Since many parameters may be considered that satisfy these conditions among all patterns, an optimization routine can be exploited for selecting these parameters consistent with the constraints[29].

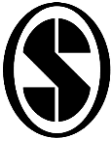

## 3. Online adaptation

The trajectories that have been generated in the last section may be realized for walking in ideal situation, where there exist neither uncertainties nor obstacles. However, for implementing the generated walking pattern on the robot in real environment, we need to employ reflexive actions as well as online modifications to tackle with uncertainties and obstacles. In this section, our main concern is to fully describe the proposed method for online adaptation in such situations. Therefore, after presenting the employed contact switch sensors, the required modifications for passing over the obstacles will be addressed.

As it is illustrated in Figure 3.a, the contact switch sensors are embedded to the corners of each foot sole. The reason for using four switches is that the contact detection may be carried out, even if the foot is not completely in contact with the ground surface. Since our main goal in this article is to deal with flat uneven terrains, there is no difference which contact switch would touch the surface. These sensors may be adjusted with respect to the foot sole to either detect the swing foot landing just on time or in advance. In order to be able to smoothly adapt the swing foot on the ground surface, the latter case is selected. Therefore, the contact switches are adjusted to detect the contact before the complete landing of the swing foot (Figure 3.b).

By introducing the offset between the swing foot sole and the ground surface at the instant of touch as $\delta_z$ (Figure 3.b), a fifth order polynomial for the swing foot trajectory in the vertical direction is considered to not only guarantee landing with no impact, but also preserve $C^2$ continuity of the modified trajectories. The boundary conditions and considered polynomial can be written as:

$$\begin{cases} \begin{cases} z_{mod}(0) = 0 \\ \dot{z}_{mod}(0) = \dot{z}_{touch\,down} \\ \ddot{z}_{mod}(0) = \ddot{z}_{touch\,down} \\ z_{mod}(T_{mod}) = -\delta_z \\ \dot{z}_{mod}(T_{mod}) = 0 \\ \ddot{z}_{mod}(T_{mod}) = 0 \end{cases} \Rightarrow z_{mod} = \sum_{i=0}^{5} a_i\, t^i \quad,\quad 0 \le t \le T_{mod} \\ z_{mod} = -\delta_z \quad\quad\quad\quad\quad\quad\quad\quad\quad\quad\quad\quad\quad\quad\quad\quad,\quad t \ge T_{mod} \end{cases} \quad (6)$$

where $z_{mod}$ is the modification that should be added to the preplanned trajectory after detecting the touch by the contact sensors. $\dot{z}_{touch\,down}$ and $\ddot{z}_{touch\,down}$ are the vertical velocity and acceleration of the swing foot at the moment of touch in the preplanned trajectory. Furthermore, $a_i$'s are coefficients that may be computed using the specified boundary conditions. Moreover, $T_{mod}$ specifies the modification time. This value may be selected such that prevent large accelerations in the period of modification which would cause large inertia forces. Hence, in our medium, $T_{mod}$ is selected as same as the time that is needed in the preplanned pattern in order for the swing foot to land on the ground surface from the distance $\delta_z$.

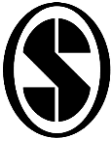

(a)            (b)

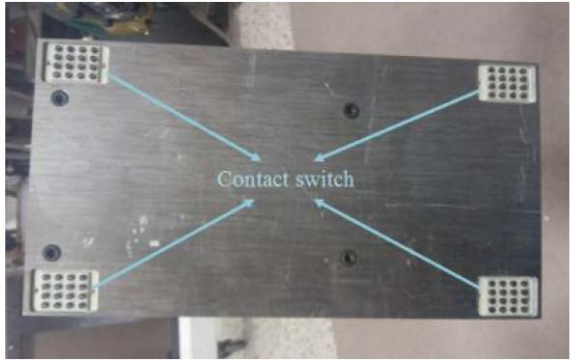
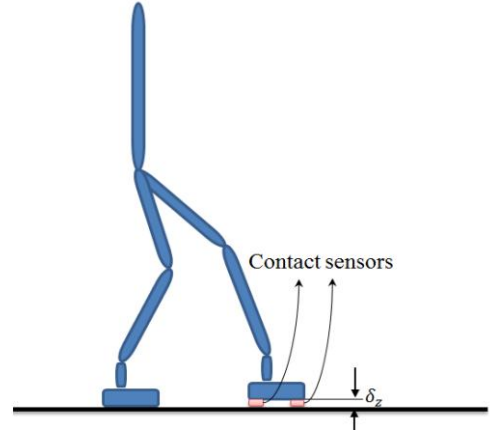

Figure 3. (a) Contact switches embedded to the foot (b) Contact switches at the instant of landing

For modification in the sagittal direction, the procedure is the same. In this direction, after detecting the touch by contact sensors, the trajectory of the swing foot should be modified in order to guarantee slippage avoidance of the swing foot. The modification time is considered as same as that of the vertical direction. In this direction, the value for the modification is arbitrary. However, in order to prevent large accelerations in the modified swing foot trajectory, the modification value in the sagittal direction ($\delta_x$) is selected as the horizontal distance in the preplanned trajectory in $T_{mod}$ before the landing time. The boundary conditions as well as the considered polynomial in the sagittal direction are considered as:

$$\begin{cases} \begin{cases} x_{mod}(0) = 0 \\ \dot{x}_{mod}(0) = \dot{x}_{touch\,down} \\ \ddot{x}_{mod}(0) = \ddot{x}_{touch\,down} \\ x_{mod}(T_{mod}) = \delta_x \\ \dot{x}_{mod}(T_{mod}) = 0 \\ \ddot{x}_{mod}(T_{mod}) = 0 \end{cases} \Rightarrow x_{mod} = \sum_{i=0}^{5} a_i t^i \,, \quad 0 \leq t \leq T_{mod} \\ \delta_x \quad\quad\quad\quad\quad\quad\quad\quad\quad\quad\quad\quad, \quad t \geq T_{mod} \end{cases} \quad (7)$$

Where $x_{mod}$ is the modification in the sagittal direction, $\dot{x}_{touch\,down}$ and $\ddot{x}_{touch\,down}$ are the velocity and the acceleration of the swing foot at the moment of touch.

The proposed adaptation algorithm is described through a block diagram and related time-tables for modifications. As it is specified in the first time-table of Figure 4, the swing foot is monitored after midpoint of the SSP (the swing foot is on its maximum height at this point) until the end of the DSP. If the swing foot lands on the ground during the SSP time period, it means that soon landing occurs. On the contrary, if the swing foot lands on the ground after the end of SSP time period, the swing foot experiences late landing. In both cases, consistent modifications are added to the preplanned trajectories in the task space to preserve smooth landing of the swing foot on the ground surface. As it is illustrated in the block diagram of Figure 4, after generating gait parameters by the gait coordinator module, consistent with the high-level walking command, the walking pattern generation is carried out based on the method that have been presented in the last section. Then, these task space trajectories are modified based on the algorithm that is exploited in online adaption module, for both advance and late landing which will be fully addressed in the next subsections.

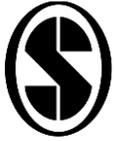

In the SURENA III humanoid robot structure, the three hip joints intersect in one point. As a result, we can employ the analytical method proposed by Park et al.[12] to compute the joint space trajectories from the task space trajectories in real-time. In this method, a reverse decoupling approach is employed to decouple the end-point position and orientation of a 6 DoF leg of a humanoid robot whose the hip joints intersect in one point. For this leg, the end-point is the foot and its desired trajectory is related with respect to the moving base of the robot (pelvis). By reversing the forward kinematics equations, we can rewrite the equations such that instead of the first three joints, the last three joints intersect in one point in the kinematic chain. Using this concept, position of the end-point in the reversed chain is independent of the last three joints. As a result, given the desired position of the end-point, a closed form solution for the first three joints is obtained. Then, using the desired orientation of the end-point frame, the trajectories for the last three joints are obtained to realize the desired orientation.

In online adaptation module of the block diagram of Figure 4, the decisions are made based on monitoring the current time and the first time-table of Figure 4. In fact, by comparing the time that the touch is detected with the time-table, it can be found whether advance landing (Figure 5.a) or retard landing (Figure 5.b) has occurred. Then, consistent with each case, suitable modification is applied to the preplanned task space trajectories that are fed to the online adaptation module, in order to yield modified task space trajectories.

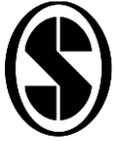
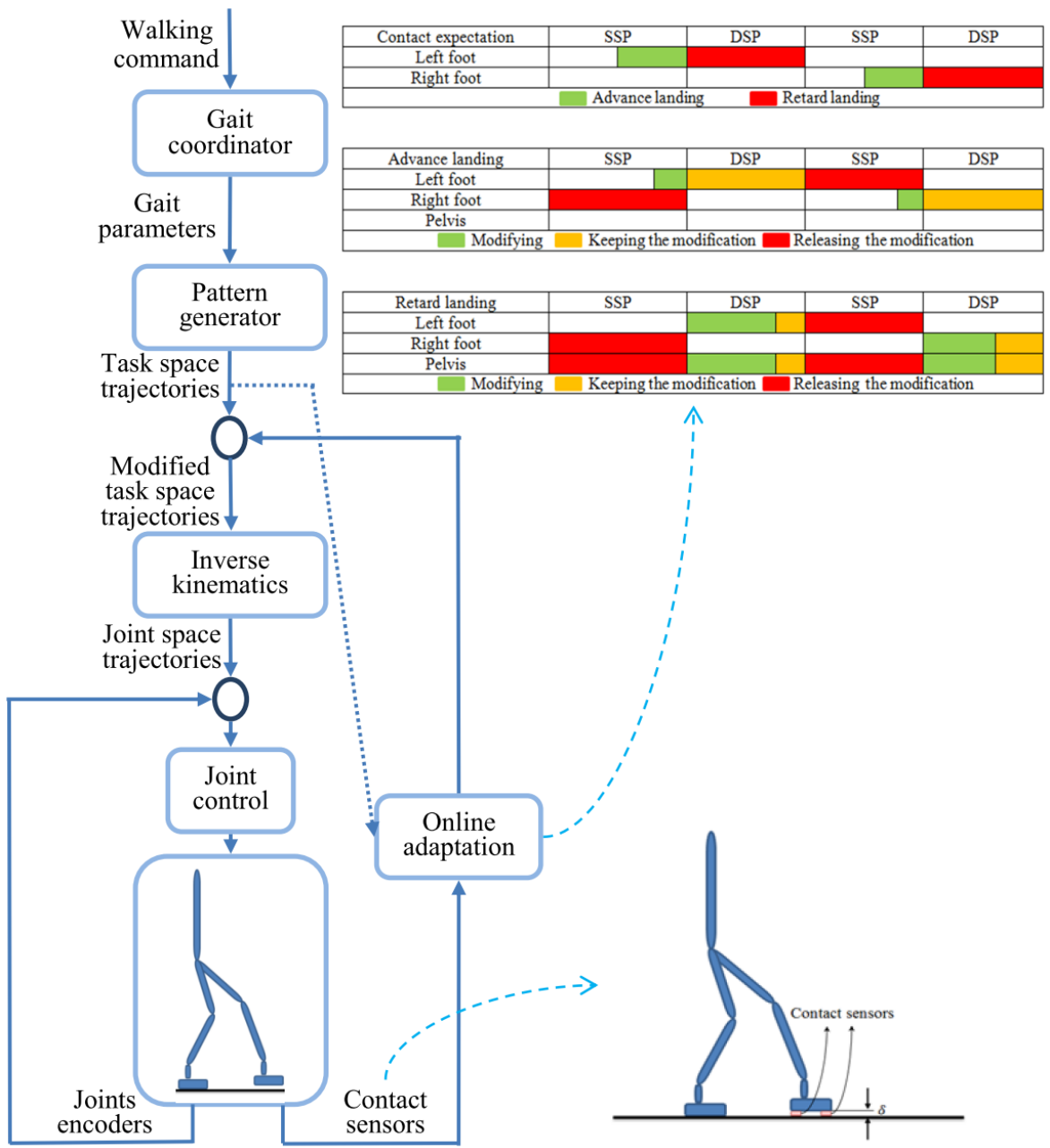

Figure 4. Block diagram of gait planning and online adaptation algorithm with related time-tables

## 3.1 Advance landing of the swing foot

There are two possibilities that may cause advancement in swing foot landing during walking. The first one is that due to deflections of the links, these values are accumulated to show their effect at the end of the kinematic tree where the swing foot is located. Furthermore, the compliance of the contact between the stance foot and the ground surface may cause swing foot advance landing. The second possibility is that an obstacle exists on the robot path or maybe the surface has some irregularities. As it is illustrated in Figure 5.a, the preplanned walking pattern should be modified in order for the generated gait to be realized. The modifications are both in the direction of motion and the vertical direction. In vertical direction, the modification should be such that preserve landing without impact. The required value for modification in each control cycle depends on the touch down height and the preplanned height at that moment:

$$\Delta z_{foot,online} = z_{touch\ down} - z_{preplanned} \quad (8)$$

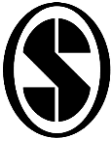

In this equation $z_{touch\ down}$ and $z_{preplanned}$ describe the height at the stance of touch and the height of swing foot in the preplanned trajectory at the current time, respectively. $\Delta z_{foot,online}$ specifies the modification that should be added to the preplanned trajectory in vertical direction to preserve smooth landing of the swing foot without impact. Hence, the modified swing foot pattern may be stated as:

$$z_f = z_{preplanned} + z_{mod} + \Delta z_{foot,online} \quad (9)$$

where $z_f$ is the modified swing foot trajectory that is exploited to yield joint space trajectories. Moreover, $z_{mod}$ may be obtained from Eq.6. Further inspections on Eq.8 and Eq.9 reveal that after detecting the touch by the contact switches, the swing foot is fixed to the $z_{touch\ down}$ and using a fifth order polynomial of Eq.6, the smooth landing would be realized.

Similarly, in the direction of motion, the modification should be such that prevent slippage of the landing foot on the new surface. By introducing the modification value in sagittal direction as:

$$\Delta x_{foot,online} = x_{touch\ down} - x_{preplanned} \quad (10)$$

And adding this modification value, as well as $x_{mod}$ which guarantee smooth landing of the swing foot (Eq.7), the modified swing foot trajectory may be specified as:

$$x_f = x_{preplanned} + x_{mod} + \Delta x_{foot,online} \quad (11)$$

In the case of advance landing of swing foot, by just modifying the swing foot trajectory, the adaptation is realized and no amendment is required for the pelvis trajectory. Hence, just the swing leg joints are modified to realize the adaptation (Figure 5.a).

## 3.2 Retard landing of the swing foot

In the case that there is a hole in the robot path, or the deflection of the links and compliance between the stance foot and the ground surface cause the swing foot lands on the ground later than it is expected, the late landing of the swing foot occurs. In this case, if no modification is applied, the robot would incline forward which can cause its fall. Under such circumstances, the modification is considered such that if the contact switches do not detect landing on the exact time, a fixed negative amount is added to the preplanned walking pattern in each time increment until the switches detect the touch:

$$\left(\Delta z_{foot,online}\right)_i = \left(\Delta z_{foot,online}\right)_{i-1} - h \quad , \quad if \ and \begin{cases} T_S \leq t \leq T_c \\ con.sw. = 0 \end{cases} \tag{12}$$

where $\left(\Delta z_{foot,online}\right)_{i-1}$ and $\left(\Delta z_{foot,online}\right)_i$ are the modification values in two successive time increments. Furthermore, $h$ is a constant amount which specifies the modification value in each time increment. This value is different for implementations with different control time cycles. Finally, the $con.sw.$ is a parameter which monitors the contact between the swing foot and the ground surface, while returns zero for the case that swing foot does not touch the ground surface. In fact, this equation implies that if the contact switches do not detect the contact after the SSP time period, the height of the swing foot is lowered by a fixed value in each increment, until the contact switches detect the touch.

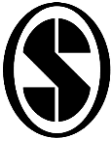

In order to obtain the modified value of swing foot trajectory in the case of retard landing, the contact state should be monitored. As long as the contact switches do not detect the touch, the swing foot height should be alleviated. Once the contact is detected, the smooth landing should be realized. Hence, the procedure is:

$$\begin{cases} z_f = z_{preplanned} + \Delta z_{foot,online} \quad , \quad if \ con.sw. = 0 \\ z_f = z_{preplanned} + z_{mod} + \Delta z_{foot,online} \ , \quad if \ con.sw. = 1 \end{cases} \tag{13}$$

In the case that the hole depth is relatively large, the modifications could cause singularity of the swing leg, due to the modifications. To tackle with this problem, the pelvis trajectory in vertical direction is modified, as well:

$$\Delta z_{pelvis,online} = \Delta z_{foot,online} \quad , \quad if \ and \begin{cases} T_S \leq t \leq T_c \\ con.sw. = 0 \end{cases} \tag{14}$$

where $\Delta z_{pelvis,online}$ is the pelvis modification in the vertical direction. By adding this value to the preplanned trajectory of the pelvis, the modified pelvis trajectory is obtained:

$$z_p = z_{preplanned} + \Delta z_{pelvis,online} \tag{15}$$

In this equation, $z_p$ specifies the modified pelvis trajectory. It should be noted that in the case of retard landing, it is not necessary to modify the swing foot trajectory in the sagittal direction. The reason is that, in this case, the velocity and acceleration of the swing foot have already reached zero, at the end of the SSP.

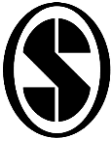

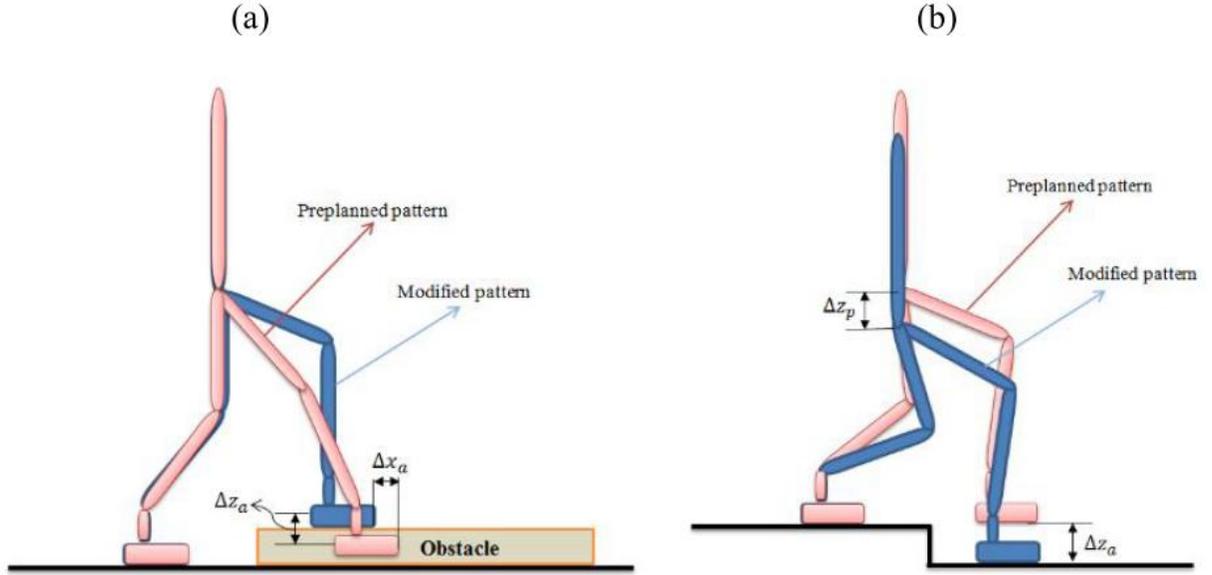

Figure 5. (a) Advance swing foot landing in the presence of a flat obstacle (b) Retard landing of swing foot in the presence of a hole

## 3.3 Releasing the modifications

The modifications that have been applied for both retard and advance landing should be released and the trajectories should become the preplanned trajectories in the absence of irregularities. The second and third time-tables of Figure 4 illustrate the procedure of applying and releasing the modifications for advance and retard landing, respectively.

For the case that the swing foot touches the ground sooner than it is planned, the modifications are applied based on Eq.11, until the end of the SSP period. Then, during the DSP, the modifications are kept to preserve the adaptation during this phase (Figure 5). Finally, during the next SSP when the modified foot became stance, the modification should be released in order for the adaptation to the new surface is done. By introducing the applied modification for each task space component as $\Delta$, this value reduces smoothly to zero, using the following trajectory:

$$\begin{cases} \Delta(0) = \Delta_{online} \\ \dot{\Delta}(0) = 0 \\ \ddot{\Delta}(0) = 0 \\ \Delta(T_s) = 0 \\ \dot{\Delta}(T_s) = 0 \\ \ddot{\Delta}(T_s) = 0 \end{cases} \Rightarrow \Delta = \sum_{i=0}^{5} a_i t^i \quad , \quad 0 \leq t \leq T_s \tag{16}$$

Similarly, for the case that the swing foot lands on the ground late, modifications should be released. In this case, as it is illustrated in third time-table of Figure 4, after the swing foot touches the ground during the DSP and the appropriate adaptation is made, the modifications are preserved in the rest of DSP. Then, during the next SSP, all the task space modification components are released, according to Eq.16, which guarantee adaptation to the new surface height.

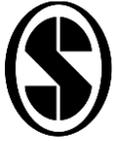

## 4. Obtained results and discussions

In this section, implementation of the proposed adaptation algorithm on the humanoid robot SURENA III will be discussed. The mass properties and geometric attributes of the considered robot are specified in Table 1. The humanoid robot SURENA III is a position-controlled robot. The lower-body joints are actuated by EC motors and a combined timing belt-pulley and harmonic drive module is exploited for power transmission. In our robot, Rigid and light-weight structure is exploited to enhance performance of our low-level control. In order to alleviate the vibrations of the joints which may exist due to flexibility of harmonic drive or the cross modules (in ankle and hip), high resolution absolute encoders at the output of the joints (load-side) are employed. Together with incremental encoders at the motor side of each joint, a cascade position control loop has been exploited. For our robot with high structural stiffness, driven by electric motors through reduction gears with relatively high ratios (gear ratio>100), very low tracking errors at the output of the joints is achieved using a simple decoupled joint position control. Furthermore, direct joint position control yields good disturbance rejection and compensation of gear friction with high bandwidth.

Table 1. Mass properties and geometric attributes of the under-study robot

| Link | Mass (gr) | Parameter | Value (mm) |
|---|---|---|---|
| Foot | 3859 | Foot length | 265 |
| Ankle | 2236 | Foot width | 160 |
| Shank | 4561 | Ankle joints height | 98 |
| Thigh | 6327 | Shank length | 360 |
| Pelvis | 17800 | Thigh length | 360 |
| Upper-body | 28482 | Distance between hip-rotation joints | 230 |
| | | Distance between hip and pelvis | 115 |
| Total Weight: 82 (Kg) | | Distance between pelvis and head | 767 |
| | | Total Height: 1.85 (m) | |

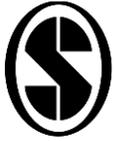

 In order to examine the effectiveness of the proposed approach, a pattern for walking on a flat surface is generated[29]. Then, during implementation of the generated walking pattern, two flat obstacles with different heights are located in the robot path. It should be noted that during stepping on the obstacles, the swing foot lands on the ground surface sooner than the considered time in the preplanned trajectory. However, when the robot descends the obstacles, the swing foot lands on the ground later than the preplanned time, while the adaptation algorithm prevents the robot from the fall due to the late landing. Therefore, during this implementation, both advance and retard landing are tested.

In Figure 6, the position components of the feet trajectories are illustrated. As it may be observed, each foot is fixed during half of the cycle and moves forward during the rest of the cycle time in sagittal direction. In the lateral direction, distance between the feet remains constant to prevent self-collision of the legs during walking. In the vertical direction, the feet move to reach their maximum height and land on the ground, and remain fixed during the rest of the gait cycle. It is noteworthy that the maximum height of the feet in walking pattern, determine maximum height of the obstacle that the robot can pass over by online adaptation. Furthermore, it should be noted that since the generated walking pattern has no heel/toe off motion, the orientation components of the feet remain fixed during walking which causes the feet to be parallel to the ground surface, so the orientation components are not illustrated in this figure.

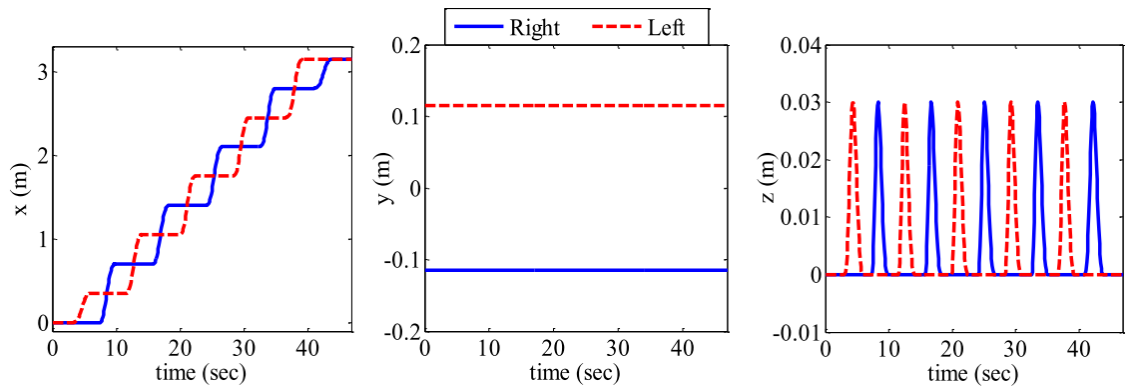

Figure 6.  Feet 3D trajectories during four walking cycles

In Figure 7, the pelvis position components of the preplanned trajectories are plotted. As it can be observed, the pelvis in the sagittal direction moves one stride in each walking cycle. Moreover, boundary conditions of the pelvis for transition between the SSP and the DSP are considered such that provide an optimal pattern[29]. In the lateral direction, a symmetric motion is considered to preserve feasibility of motion in this direction. Finally, in the vertical direction, the motion is employed to avoid singularity of the knee during walking. Furthermore, the orientation components of the pelvis are considered to be fixed during walking, so are not shown in this figure.

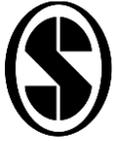

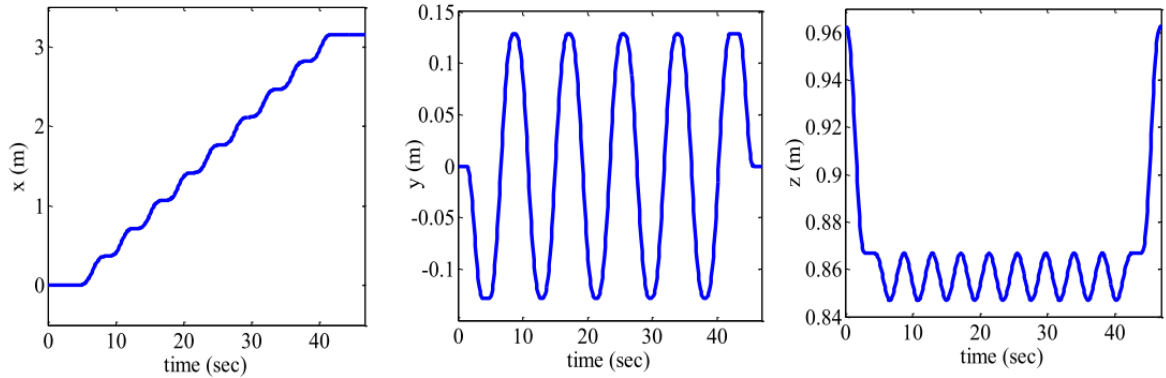

Figure 7. Pelvis 3D trajectories during four walking cycles

In order to demonstrate that the generated walking pattern in the absence of obstacles is feasible, the ZMP trajectory in both sagittal and lateral directions are illustrated in Figure 8 and Figure 9. As it may be observed in these figures, the ZMP remains inside the support polygon during walking which means the dynamic balance is guaranteed. Furthermore, it is noteworthy that the ZMP trajectory is continuous during walking, thanks to the $C^2$ continuity of the generated trajectories in the task space.

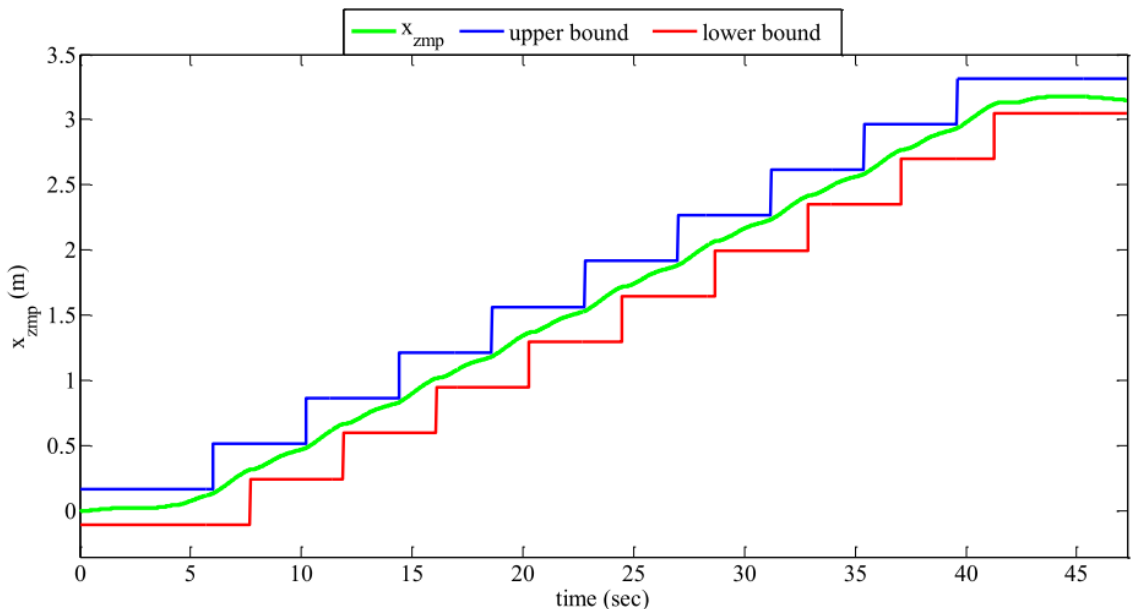

Figure 8. ZMP trajectory in the sagittal direction during four walking cycles

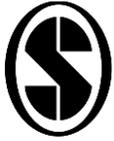

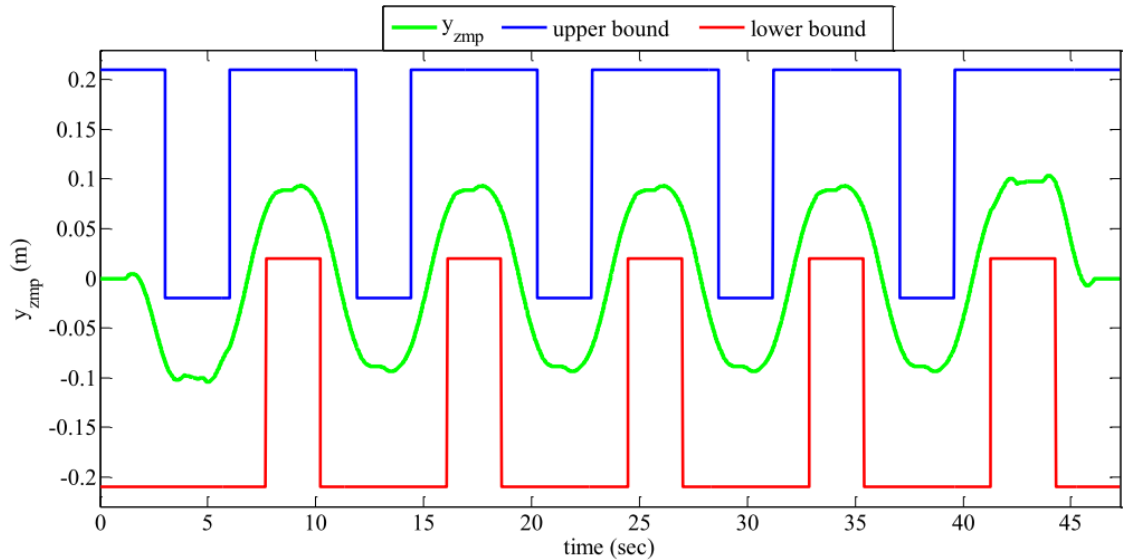

Figure 9. ZMP trajectory in the lateral direction during four walking cycles

In order to investigate performance of the proposed online adaptation algorithm, the generated walking pattern is implemented on the real robot. During walking, three obstacles are put in the robot path. No information about the heights and the surface characterizations of the obstacles has been considered in the gait planning procedure, as it can be seen in Figure 6 and Figure 7. In Figure 10, demonstration of the realized walking pattern in the presence of the obstacles is shown. As it can be observed in these snapshots, the robot walks over the obstacle during one step, and descends the obstacle during the next step.

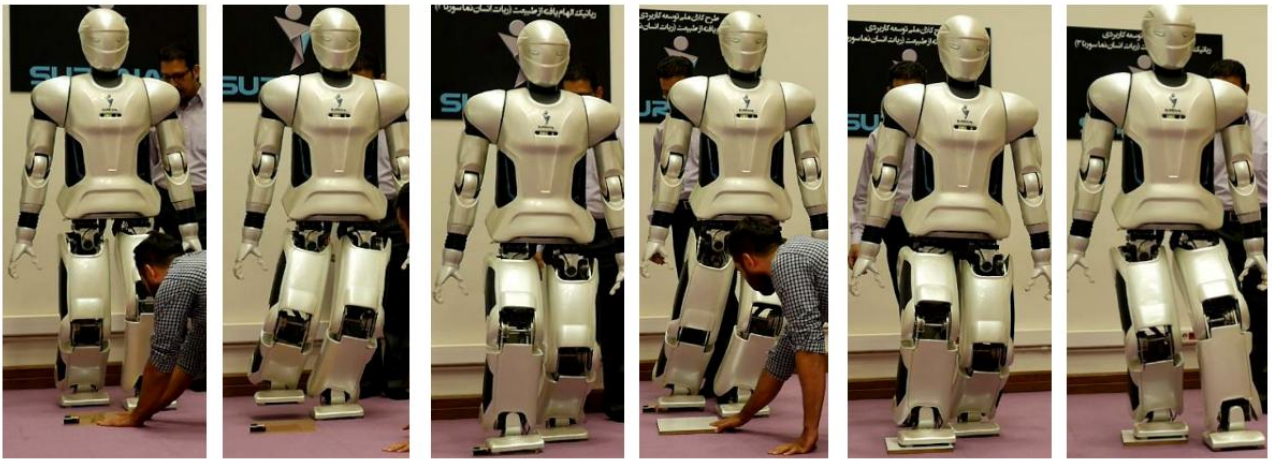

Figure 10. snapshots of walking on a surface with unknown obstacles using SURENA III, a humanoid robot designed and fabricated in CAST, University of Tehran,

In Figure 11, the online modifications that are applied to the feet trajectories are shown. During walking, three obstacles are located to the robot path, while the exact moments that the sensors detect the obstacles are specified in this figure. As it may be observed, after detecting the first obstacle, the modifications are applied to the swing foot trajectory until the end of the SSP time period. It can be seen that the modification in the vertical direction is positive, which means the swing foot lands on the ground sooner than it is expected, so the swing foot new reference should be upper than the preplanned one to prevent impact. However, the modification in the sagittal direction is negative, which shows that the swing foot is kept behind the preplanned trajectory to be adapted to the new surface without slippage. During the DSP period, the modifications are preserved. Then, during the next SSP, where the foot whose trajectories have been modified is stance, the modifications are released.

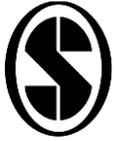

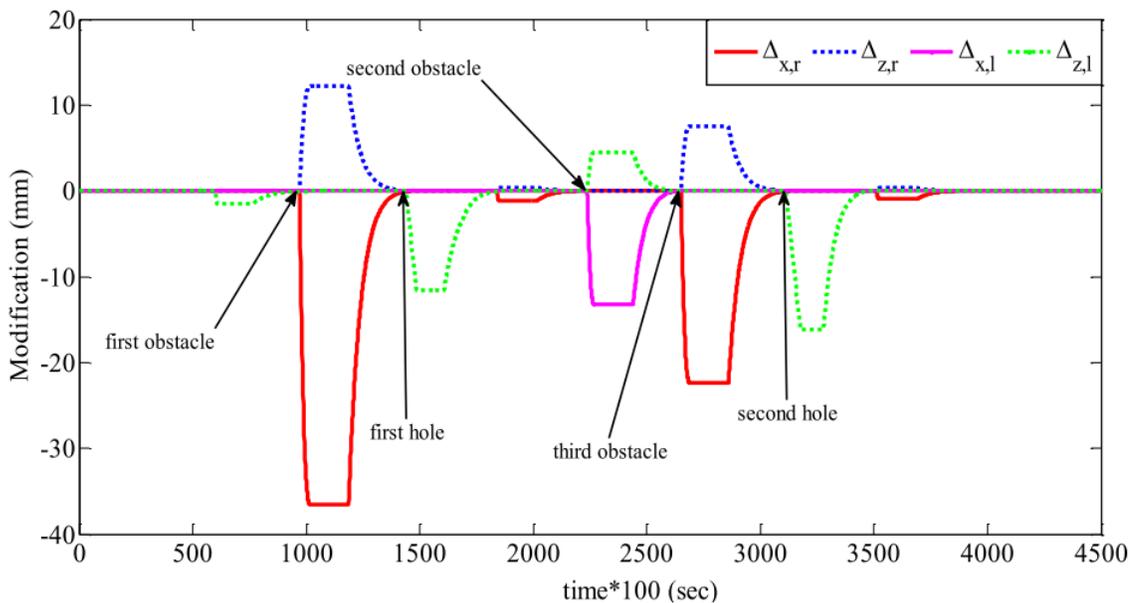

Figure 11. The applied modifications in task space

The releasing period is from the initiation of the SSP to the termination of the SSP, which realizes adaptation to the new surface height. After passing over the obstacle and adaptation to the new surface height, the next swing foot is prone to land on the ground surface late. Therefore, as it is specified in Figure 11, this situation resembles a hole in the robot path. Under such circumstances, the swing foot lowers its height, until the contact sensors detect the touch. In this case, the modification in the vertical direction is negative, while there is no modification applied to the sagittal direction. Moreover, the same modification in vertical direction for the pelvis trajectory is applied to preserve singularity avoidance. After detection of contact, the modification in the vertical direction is kept during the rest of the DSP period. Finally, during the next SSP, this modification is released.

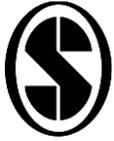

Since in each walking gait, the contact state is monitored and the adaptation is applied, some slight modifications may be observed in Figure 11, while there is neither obstacle nor hole in the robot path. This means the swing foot is prone to land on the ground sooner or later than expected, due to uncertainties and surface irregularities. Therefore, the proposed algorithm is suitable for walking on the surfaces without obstacles, while the robot links deflections as well as compliance between the stance foot and the ground surface may cause retard or advance landing of the swing foot.

Finally, in order to demonstrate that the robot feet do not impact the ground using the proposed adaptation algorithm, the vertical forces measured by the force/torque sensors which are embedded to the ankles are shown in Figure 12. In this figure, the instances that the robot feet touch the obstacle are specified. As it can be observed in this figure, no impacts may be observed at the moments that the robot steps on the obstacles. The reason that the obtained data are not filtered is that filtering may cause elimination of the impacts due to their high frequency behavior. Therefore, in this figure, the raw data from the force/ torque sensors without filtering are plotted.

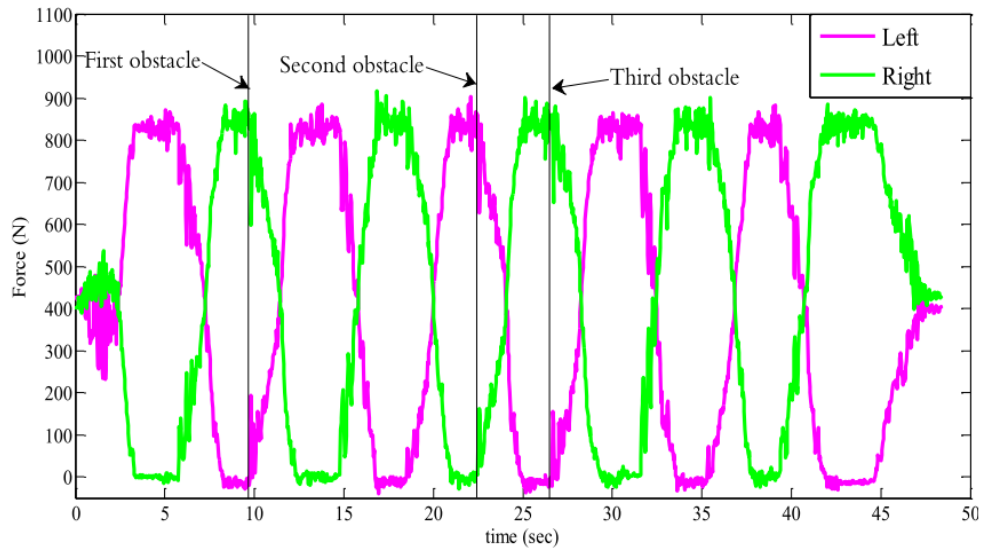

Figure 12. Vertical interaction force between the feet and the ground surface in the presence of unknown obstacles which reveals no impact

Furthermore, the obtained data from the IMU in both lateral and sagittal directions show that walking over obstacles does not affect the orientation of the upper-body (Figure 13). In this case, again the raw data obtained from IMU is used to plot the time history of the upper-body orientation.

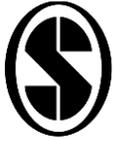

As it may be observed in this figure, no violation may be seen on the upper-body orientation at the instances that the swing foot touches the obstacle.

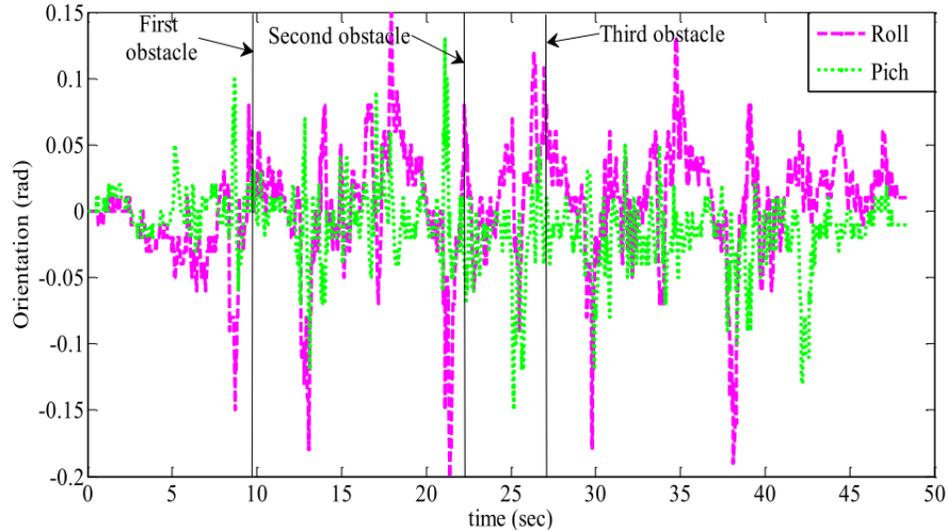

Figure 13. Orientation of the upper-body during walking

## 5. Conclusions

In this study, a method for online adaptation of humanoid robots walking on uneven terrains has been suggested. In this method, the contact switch sensors have been exploited to figure out that the swing foot has a certain distance to the ground surface. Knowing the distance, the preplanned trajectories in the task space have been modified to realize landing without impact or slippage. The implementation of the proposed approach on the SURENA III humanoid robot, walking in the presence of flat obstacles, revealed that this strategy can deal with both advance and retard landing of the swing foot. Also, the information that have been obtained from the force/torque sensors embedded to the ankles and the IMU mounted on the upper-body demonstrated that by adopting the suggested online algorithm, neither the swing foot impacts the ground surface nor the upper-body orientation is violated.

**Acknowledgment** The authors would like to express deep gratitude to the Industrial Development and Renovation Organization of Iran (IDRO) and Iran National Science Foundation (INSF) for their financial support (Project Number: 94000927) to develop the SURENA III humanoid robot. We further thank to the members of CAST for their valuable participation in the design and fabrication of the robot.

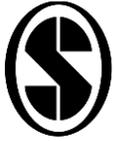

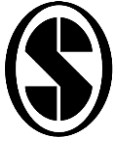